\documentclass[runningheads]{llncs}
\usepackage[T1]{fontenc}
\usepackage{booktabs}
\usepackage{hyperref}
\usepackage{graphicx}
\usepackage{textgreek}
\begin{document}
\title{When CQs Go Wrong: Challenges in CQ Verification with OE-Assist}
\def\thefootnote{*}\footnotetext{Equal contribution.}
\titlerunning{When CQs go wrong}
%
\author{Anna Sofia Lippolis\inst{1,2,*} \and
Mohammad Javad Saeedizade\inst{3,*} \and
Eva Blomqvist\inst{3}\and
Andrea Giovanni Nuzzolese\inst{2}\and
Aldo Gangemi\inst{1,2}\and
Robin Keskisärkkä \inst{3}
}
\authorrunning{A. Lippolis et al.}
%
\institute{University of Bologna, Italy\\
\email{\{annasofia.lippolis2,aldo.gangemi\}@unibo.it}
\and
ISTC-CNR, Italy\\
\email{andrea.nuzzolese@istc.cnr.it}
\and
Linköping University, Sweden\\
\email{\{Javad.saeedizade,eva.blomqvist,robin.keskisarkka\}@liu.se}}
\maketitle              
\begin{abstract}
Competency Questions (CQs) are the central component of CQ-verification, an established process in which an ontology is evaluated against a set of natural language questions to determine whether the intended purpose of the ontology has been properly modelled. However, CQ-verification is often time-consuming and error-prone, as it requires careful interpretation of linguistic nuances and precise alignment with formal ontology constructs. Ambiguities and complexity in CQs can further complicate this process, leading to inconsistent modelling decisions and verification outcomes.
In this paper, we investigate what makes a CQ challenging and possible solutions to enhance the users' performance in the CQ-verification process. We experimented with the data of 19 participants who performed CQ-verification on 20 tasks using an LLM assistant to support ontology evaluation. The results show the necessity of a tool to refine CQs before publishing them to avoid ambiguity or excessive complexity in later phases of the ontology engineering process.

\keywords{Ontology Engineering \and LLMs \and Competency Questions}
\end{abstract}
\section{Introduction and related work}

Competency Questions (CQs), and in particular, verification CQs~\cite{keet2024discerning}, are fundamental tools in knowledge engineering. They are requirements that an ontology must be able to answer. In fact, the long-standing practice of CQ-verification~\cite{blomqvist2012ontology} evaluates a produced ontology based on whether it correctly models the knowledge necessary to answer a predefined set of CQs. Although conceptually straightforward, this process is time-consuming and prone to interpretation errors, which can derive from the set of CQs itself. Indeed, as noted in many studies, quality dimensions such as ambiguity, readability, and complexity in CQs~\cite{alharbi2024review,alharbi2025comparative} could result in inconsistent modelling decisions and verification outcomes, and even carefully curated CQ benchmarks are not immune to such issues. Still, very little guidance exists for formulating CQs and assessing whether they constitute good CQs~\cite{keet2024discerning,keet2024roles}. At the same time, the increasing capabilities of LLMs have introduced new possibilities for automating and assisting CQ-related tasks from a variety of scenarios, including CQ generation, refinement, formalisation, and ontology generation and evaluation~\cite{lippolis2025bench4ke,lippolis2025ontology,saeedizade2025large,saeedizade2024navigating}. As ontology development becomes more automated, understanding how ambiguity in CQs affects ontology development and verification performance becomes increasingly important.

In this work, we empirically investigate ambiguity and complexity in CQs within the context of ontology evaluation. We conducted a controlled CQ-verification experiment with 19 ontology engineers in OE-Assist~\cite{lippolis2025large}, whose resources are available on \href{https://github.com/dersuchendee/OE-Assist}{Github}, and we analysed free-text user comments from that experiment to study more variables affecting users' performance. Our main research question is ``How can CQ formulations be evaluated to identify the features that make them challenging in CQ-verification?'' By examining users' feedback and the time spent on each CQ w.r.t the length of the ontology, CQ complexity, and the time they spent on the experiment, we aim to provide empirical insights that inform ontology engineering practices to formulate CQs.
The final aim of this work is to explore pitfalls that occur during the CQ creation process and to investigate how such issues might be identified, potentially informing future tool support.
\vspace{-.5em}



\section{Experimental setup}
This section summarises the semi-automatic, human-in-the-loop component of the OE-Assist study and then outlines the additional factors not examined in the original work, together with the empirical approach adopted here.

Lippolis et al.~\cite{lippolis2025large} presented OE-Assist, a prototype designed to support in a semi-automatic way the CQ-verification workflow of Blomqvist et al.~\cite{blomqvist2012ontology}. In CQ-verification, ontology engineers determine whether an ontology captures the information required to answer a natural-language CQ. Operationally, a CQ is considered modelled if a SPARQL test query can be built that returns the expected answer; otherwise it is not. The key elements of the OE-Assist experiment are as follows. \textbf{Participants:} $N = 19$ ontology engineers from academic networks with varying levels of expertise. \textbf{Conditions:} Each participant completed verification tasks under two settings: \emph{assisted} (LLM-generated suggestions available) and \emph{unassisted} (Protégé only). Each participant indicated whether a CQs was modelled or not and the level of difficulty on a Likert scale from 1 to 5. We also recorded qualitative feedback in free text. \textbf{Outcome of OE-Assist:} The results indicate strong reliance on LLM suggestions: correct suggestions improved performance, whereas incorrect or misleading suggestions reduced performance relative to the unassisted condition.

Here, we investigate how characteristics of CQ formulation and user stories relate to human verification performance. Our objectives are (1) to identify CQ- and ontology-level factors associated with decision accuracy, perceived difficulty and response time, and (2) to derive actionable recommendations for a future tool that serves as a CQ pitfall scanner. Concretely, we analyse the following variables extracted from the OE-Assist session notes: \textbf{User correctness:} Denoting whether the participant's CQ-verification decision (``CQ modelled'' vs.\ ``not modelled'') matches the ground truth. \textbf{Perceived Difficulty:} A score from 1 (=not difficult at all)-5 (=very difficult) given by users for a CQ-verification task. \textbf{Decision duration:} Time the participant spent on a given CQ before submitting a decision. \textbf{Ontology size:} An ordinal descriptor of the OWL artefacts used in the study (``Small'', ``Medium'', ``Large'') derived from axiom count. \textbf{CQ complexity:} A readability score computed from the CQ text. Following~\cite{alharbi2025comparative}, we compute Flesch–Kincaid Grade Level and the Gunning Fog Index and treat these indices as continuous predictors of CQ complexity.

Our analysis proceeds in two phases. We ran pairwise correlation tests (Spearman's $ρ$ and Kendall's $τ$) and appropriate pairwise comparisons to identify predictors of user correctness. Second, we perform a qualitative analysis of the session notes and feedback from the users.

\vspace{-.5em}

\section{Results}
\vspace{-.5em}

\begin{figure}
    \centering
    \includegraphics[width=.95\linewidth]{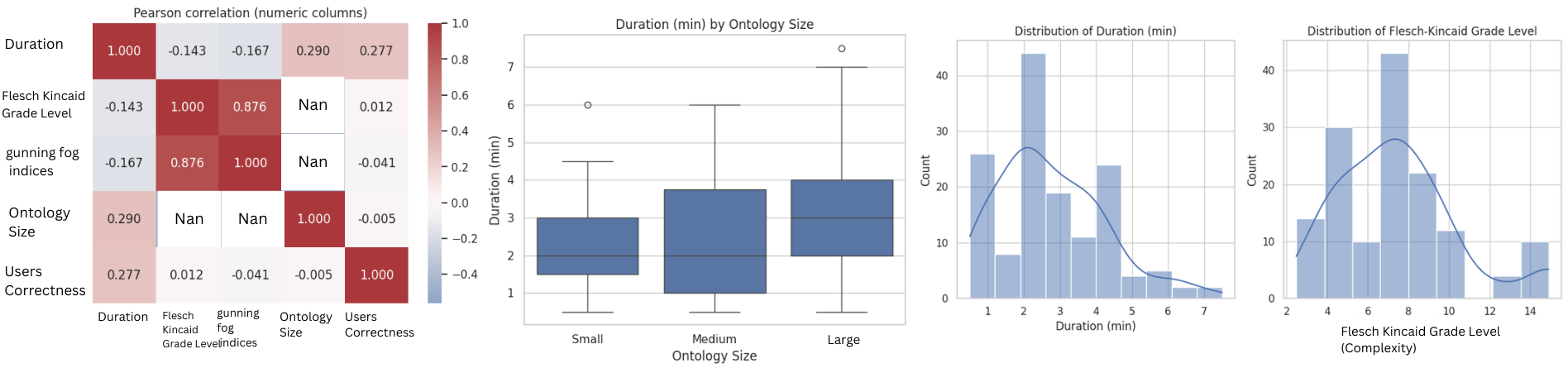}

    \label{fig:fig1}
    
    \caption{Correlation analysis of CQ completion time, user correctness, ontology size, and the two complexity metrics introduced by Alharabi et al.~\cite{alharbi2025comparative}. The figure additionally includes a box plot of completion time stratified by ontology size and statistics of time spent on each CQ duration and CQ complexity.}

\end{figure}



The results indicate that only one statistically significant association was observed. Specifically, a moderate positive correlation was found between decision duration and perceived difficulty (Spearman's $\rho = 0.42$; Kendall's $\tau = 0.34$; both $p < 0.001$), suggesting that tasks requiring longer completion times were more likely to be rated as difficult by users.

No statistically significant associations were identified among the other examined variable pairs, including decision duration and CQ complexity, decision duration and ontology size, or CQ complexity and ontology size. These findings suggest that, within the present dataset, perceived difficulty is primarily related to time spent on the task.


\begin{table}[h]
\centering
\caption{Semantic range of the word ``resource'' and some European equivalents. }

\resizebox{0.7\linewidth}{!}{%
\begin{tabular}{lll}
\toprule

\textbf{Language} & \textbf{ Nearest word} & \textbf{     Expanded meaning in English} \\
\midrule
English    
& resource  
& supply; asset; means; source; natural resource  \\

French     
& ressource 
& resource; supply; \textbf{financial means}; personal capability \\

German     
& Ressource 
& resource; raw material; energy supply; 
  \textbf{economic asset} \\

Spanish    
& recurso   
& resource; means; \textbf{financial means}; \textbf{legal appeal}; rhetorical device\\

Portuguese 
& recurso   
& resource; means; \textbf{financial means}; legal appeal \\

Italian    
& risorsa   
& resource; asset; personnel; \textbf{financial means} \\

Swedish    
& resurs    
& resource; asset; funding; personnel allocation; \\
\bottomrule
\end{tabular}}
\end{table}


However, qualitative feedback collected during the experiment revealed three major issues that affected participants’ perceived difficulty and accuracy:
(1) In some cases, the SPARQL queries automatically generated from the complex CQs appeared syntactically incorrect or structurally unusual.
(2) Certain CQs were perceived as complex/difficult to understand.
(3) Certain CQs were repeatedly described as ambiguous.

With respect to (1), improvements in SPARQL query generation, particularly through more robust and up-to-date generation techniques, could enhance system reliability and overall user experience.
Regarding (2), only four CQs exhibited a complexity score (Flesch-Kincaid Grade Level) above 10
    (scores ranged from 2 to 14).
Notably, all the four CQs were explicitly identified by participants as difficult to comprehend. Thus, readability metrics are useful to anticipate difficulties and should be considered during CQ formulation.
Concerning (3), as an example, users highlighted the question ``Which resources mention an organ builder?'' as an ambiguous CQ. Specifically, the term ``resource'' was reported as unclear, partly because it resembles semantically different words in several European languages. 
Cognates of ``resource'' across languages may carry additional meanings (e.g., financial means, legal appeal) in German, French, Spanish, Italian, and Portuguese, which can influence interpretation. In this case, some users searched for financial elements within the ontology when attempting to answer the question. A potential solution is a tool that assesses CQ wording using LLMs;
    for example, it could suggest a clarification such as:
``Which resources (e.g., documents or records) mention an organ builder?''.

\section{Conclusion}

    In this work, we
examined additional variables in the OE-Assist study to identify factors influencing CQ-verification performance beyond LLM accuracy. The findings indicate that the Flesch-Kincaid Grade Level is a useful indicator of CQ complexity, aligned with human feedback, and can help anticipate comprehension difficulties. The results also highlight the importance of lexical clarity when writing CQs. Cross-linguistic semantic interference, particularly in multilingual settings, may affect interpretation and task performance. Providing explicit conceptual clarification within CQs can therefore reduce ambiguity.
Future work will develop a CQ pitfall scanner that automatically flags unclear or complex wording and suggests fixes—analogous to existing Ontology pitfall scanner tools.

%
%
%
\bibliographystyle{splncs04}
\bibliography{bib}





\end{document}